\documentclass[10pt,twocolumn]{article}

\usepackage[a4paper,margin=1.8cm]{geometry}
\usepackage{graphicx}
\usepackage{amsmath,amssymb}
\usepackage{booktabs}
\usepackage{multirow}
\usepackage{array}
\usepackage{caption}
\usepackage{float}
\usepackage{url}
\usepackage{hyperref}
\usepackage{enumitem}
\usepackage{titlesec}
\usepackage{xcolor}
\usepackage{pdflscape}
\usepackage{makecell}
\hypersetup{
    colorlinks=true,
    linkcolor=black,
    citecolor=black,
    urlcolor=blue
}

\titleformat{\section}{\bfseries\normalsize}{\thesection.}{0.5em}{}
\titleformat{\subsection}{\bfseries\normalsize}{\thesubsection}{0.5em}{}

\title{\textbf{AOP-Smart: A RAG-Enhanced Large Language Model Framework for Adverse Outcome Pathway Analysis}}

\author{
Qinjiang Niu$^{1}$ \\
Lu Yan$^{1,*}$ \\
\\
$^{1}$Nanyang Normal University, Nanyang, China \\
*Corresponding author: Lu Yan
}

\date{}

\begin{document}

\twocolumn[
\maketitle
\begin{abstract}
Adverse Outcome Pathways (AOPs) are an important knowledge framework in toxicological research and risk assessment. In recent years, large language models (LLMs) have gradually been applied to AOP-related question answering and mechanistic reasoning tasks. However, due to the existence of the hallucination problem, that is, the model may generate content that is inconsistent with facts or lacks evidence, their reliability is still limited.

To address this issue, this study proposes an AOP-oriented Retrieval-Augmented Generation (RAG) framework, AOP-Smart. Based on the official XML data from AOP-Wiki, this method uses Key Events (KEs), Key Event Relationships (KERs), and specific AOP information to retrieve relevant knowledge for user questions, thereby improving the reliability of the generated results of large language models.

To evaluate the effectiveness of the proposed method, this study constructed a test set containing 20 AOP-related question answering tasks, covering KE identification, upstream and downstream KE retrieval, and complex AOP retrieval tasks. Experiments were conducted on three mainstream large language models, Gemini, DeepSeek, and ChatGPT, and comparative tests were performed under two settings: without RAG and with RAG.

The experimental results show that, without using RAG, the accuracies of GPT, DeepSeek, and Gemini were 15.0\%, 35.0\%, and 20.0\%, respectively; after using RAG, their accuracies increased to 95.0\%, 100.0\%, and 95.0\%, respectively. The results indicate that AOP-Smart can significantly alleviate the hallucination problem of large language models in AOP knowledge tasks, and greatly improve the accuracy and consistency of their answers.
\end{abstract}

\vspace{0.5cm}
]

\section{Introduction}

Adverse Outcome Pathways (AOPs) are a conceptual framework used to describe the causal relationship from Molecular Initiating Events (MIEs) to Adverse Outcomes (AOs), and are of great significance in toxicological research and chemical risk assessment \cite{Ankley_2009}. By organizing Key Events (KEs) at the molecular, cellular, tissue, organ, and individual levels, AOPs provide systematic support for the expression of toxicity mechanisms, knowledge integration, and risk judgment \cite{Bal_Price_2017}. With the continuous development of AOP-Wiki, the number of related knowledge entries has been increasing, and the content covers more species, biological levels, and complex Key Event Relationships (KERs). Although the expansion of knowledge scale improves the richness of AOP resources, it also increases the difficulty for researchers in retrieving, understanding, and integrating information.

To improve the usability of AOP knowledge, previous studies have developed various supporting tools, such as AOP-HelpFinder and AOP-Explore, to support the mining, retrieval, and visualization analysis of AOP-related information \cite{Jornod_2021,Kumar_2024}. At the same time, large language models (LLMs) have shown strong capabilities in tasks such as natural language understanding, knowledge question answering, and complex reasoning, and have demonstrated broad application prospects in biomedical text mining and mechanistic knowledge analysis \cite{vaswani2017attention,brown2020language,Lee_2019}. Existing studies have shown that LLMs can assist in the extraction, integration, and structured representation of scientific knowledge, and can help with the construction of AOP chains to a certain extent \cite{Shi_2024}. However, the generated results of LLMs are limited by the coverage of training corpora and the generation mechanism itself, and they are prone to producing hallucinations when dealing with highly specialized domain knowledge, that is, outputs that are fluent in expression but factually incorrect, thereby affecting their reliability and interpretability in scientific research \cite{Ji_2023,lin2022truthfulqa}.

Retrieval-Augmented Generation (RAG) provides an effective idea for alleviating the above problems. By introducing external knowledge before LLMs generate content, RAG provides knowledge relevant to the question as context to the model, thereby enhancing the factual basis of model responses and reducing the risk of hallucination \cite{lewis2020retrieval}. A typical RAG framework usually includes a retrieval module and a generation module: the former obtains relevant content from external knowledge sources, and the latter generates answers based on the retrieved results \cite{Chen_2017,karpukhin2020dense}. Most existing RAG methods are based on vector databases and semantic similarity retrieval, and have achieved good results in open-domain question answering and knowledge-intensive tasks.

However, research on RAG in the AOP domain is still relatively limited. AOP knowledge is different from general unstructured text; it has significant hierarchical structure, causal relationships, and network characteristics, which place higher requirements on knowledge organization and retrieval methods. Existing studies still lack RAG designs oriented to the causal relationship characteristics of AOPs, especially in how to effectively utilize KEs, KERs, and overall AOP structural information to support complex AOP mechanistic knowledge modeling and reasoning.

Based on this, this paper proposes an AOP analysis-oriented RAG framework, AOP-Smart. This method uses the XML data provided by AOP-Wiki to construct a knowledge base and index files, and designs an indexing and knowledge expansion mechanism based on question-related KEs to achieve joint retrieval and knowledge integration of KE, KER, and AOP information. On this basis, the KE, KER, and AOP knowledge related to the question are used as external context and input into LLMs, thereby improving the reliability of LLM results.

The main contributions of this study are as follows:

\begin{enumerate}[label=(\arabic*)]
    \item A RAG framework for the AOP domain is proposed, which introduces AOP knowledge into the reasoning process of LLMs;
    \item A knowledge expansion strategy based on question-related KEs is designed to achieve associated retrieval and knowledge supplementation among KEs, KERs, and AOPs;
    \item Comparative experiments verify the effectiveness of this method in alleviating the hallucination problem in the AOP domain and improving answer accuracy.
\end{enumerate}

\section{Methods}

\subsection{Overall Framework of AOP-Smart}

To ensure the consistency of data sources and the reproducibility of the experimental process, this study conducted experiments based on the stable XML snapshot of the AOP-Wiki released on 2026-01-01. This dataset was obtained from the official AOP-Wiki quarterly “Permanent Downloads” archive (https://aopwiki.org/downloads
), which provides versioned XML dumps for citation and reproducible research. The 2026-01-01 release was selected to guarantee data stability, and all contents were subsequently parsed and processed in a unified manner.

It should be noted that the constructed system does not depend on a specific version of the data, but has good scalability and updating capability. By replacing the XML file of the corresponding version, the knowledge base can be updated.

\begin{figure*}[t]
    \centering
    \includegraphics[width=0.95\textwidth]{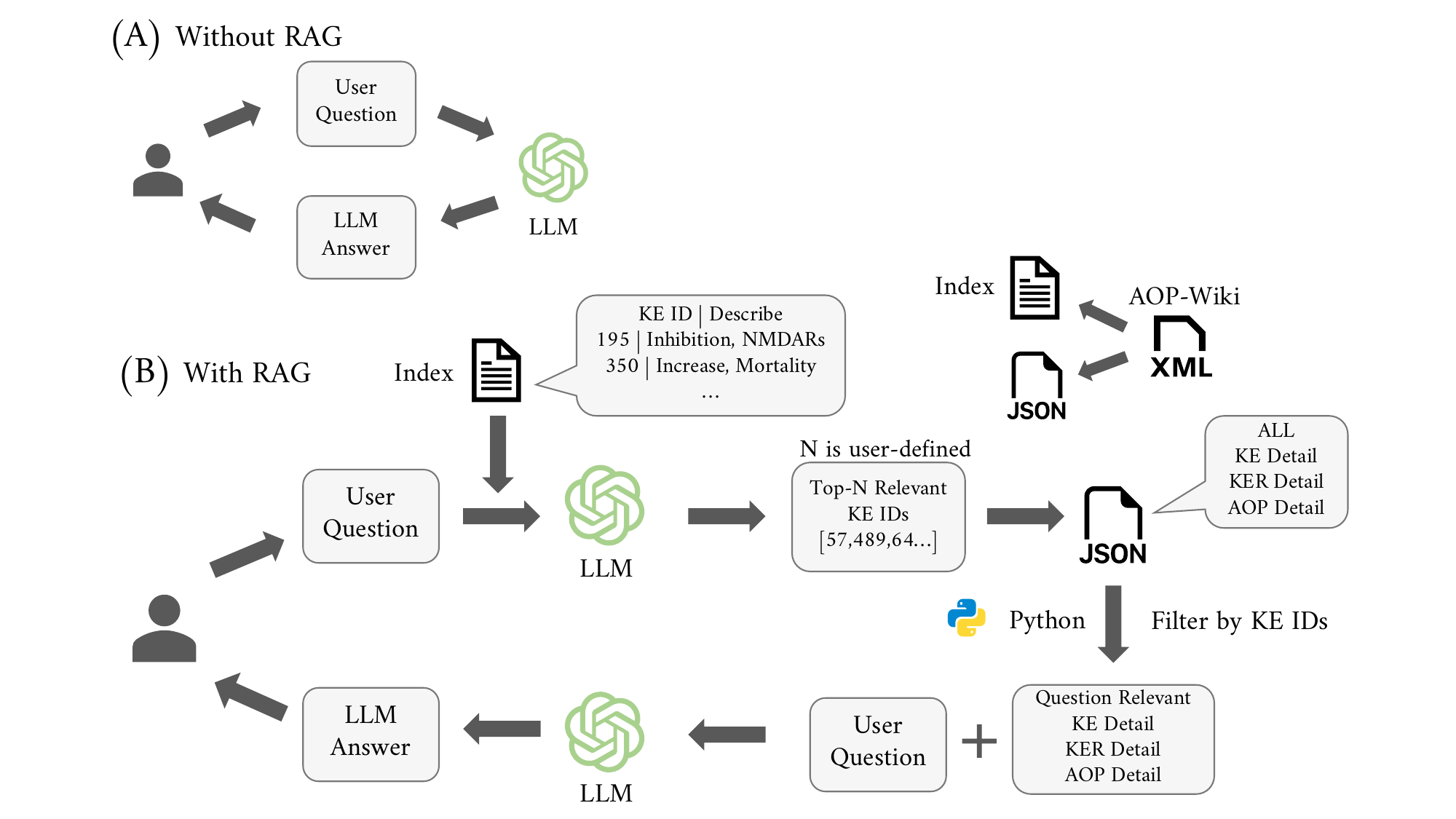}
    \caption{Overview of the AOP-Smart framework. (A) In the absence of retrieval augmentation, the LLM directly generates responses based solely on its internal parameters, which may lead to incomplete or hallucinated outputs. (B) With the proposed AOP-Smart framework, user queries are first used to retrieve relevant KE candidates from an indexed KE list, followed by structured expansion to KE, KER, and AOP-level information from the knowledge base. The retrieved structured context is then combined with the user query and fed into the LLM to produce a knowledge-enhanced and more reliable response.}
    \label{fig:framework}
\end{figure*}

As shown in Figure~\ref{fig:framework}(A), without introducing the RAG mechanism, the user question is directly input into the LLM, and the model answers only relying on its internal parameters, lacking the support of external knowledge, and thus is prone to problems such as missing information or factual errors.

As shown in Figure~\ref{fig:framework}(B), the AOP-Smart framework introduces a knowledge retrieval process prior to response generation. Specifically, the official XML data from AOP-Wiki is first parsed into two structured resources: \texttt{Index.txt} and \texttt{AOP-Smart.json}. Among them, \texttt{Index.txt} stores the IDs and titles of all key events (KEs) and is used for efficient preliminary candidate retrieval. This index is extremely lightweight, containing only approximately 20,000 tokens, which is negligible compared to the context window of modern large language models (which can reach nearly one million tokens). Therefore, it can be directly loaded into the prompt without introducing significant computational overhead, while also enabling easy extensibility for future AOP-Wiki updates. In contrast, \texttt{AOP-Smart.json} stores detailed structured information about KEs, key event relationships (KERs), and adverse outcome pathways (AOPs), supporting downstream reasoning and generation tasks.

During the inference stage, the system first inputs the user question together with the content in \texttt{Index.txt} into the LLM, and ranks the relevance of all KEs through prompt engineering, from which the Top-N most relevant KEs are selected (N is a tunable parameter). Subsequently, based on the IDs of the selected KEs, the corresponding KE details are retrieved from \texttt{AOP-Smart.json}, and their related KER and AOP information are further expanded to construct a structured context representation (including KE Detail, KER Detail, and AOP Detail).

Finally, this structured context and the user question are jointly input into the LLM for generative reasoning, thus completing the closed-loop process from question-driven retrieval to knowledge-enhanced generation. This mechanism effectively alleviates the knowledge deficiency problem of traditional LLMs in AOP tasks and significantly improves the accuracy and interpretability of the generated results.

\subsection{Expansion Method Based on TOP-N KE IDs}

\begin{figure}[t]
    \centering
    \includegraphics[width=0.9\columnwidth]{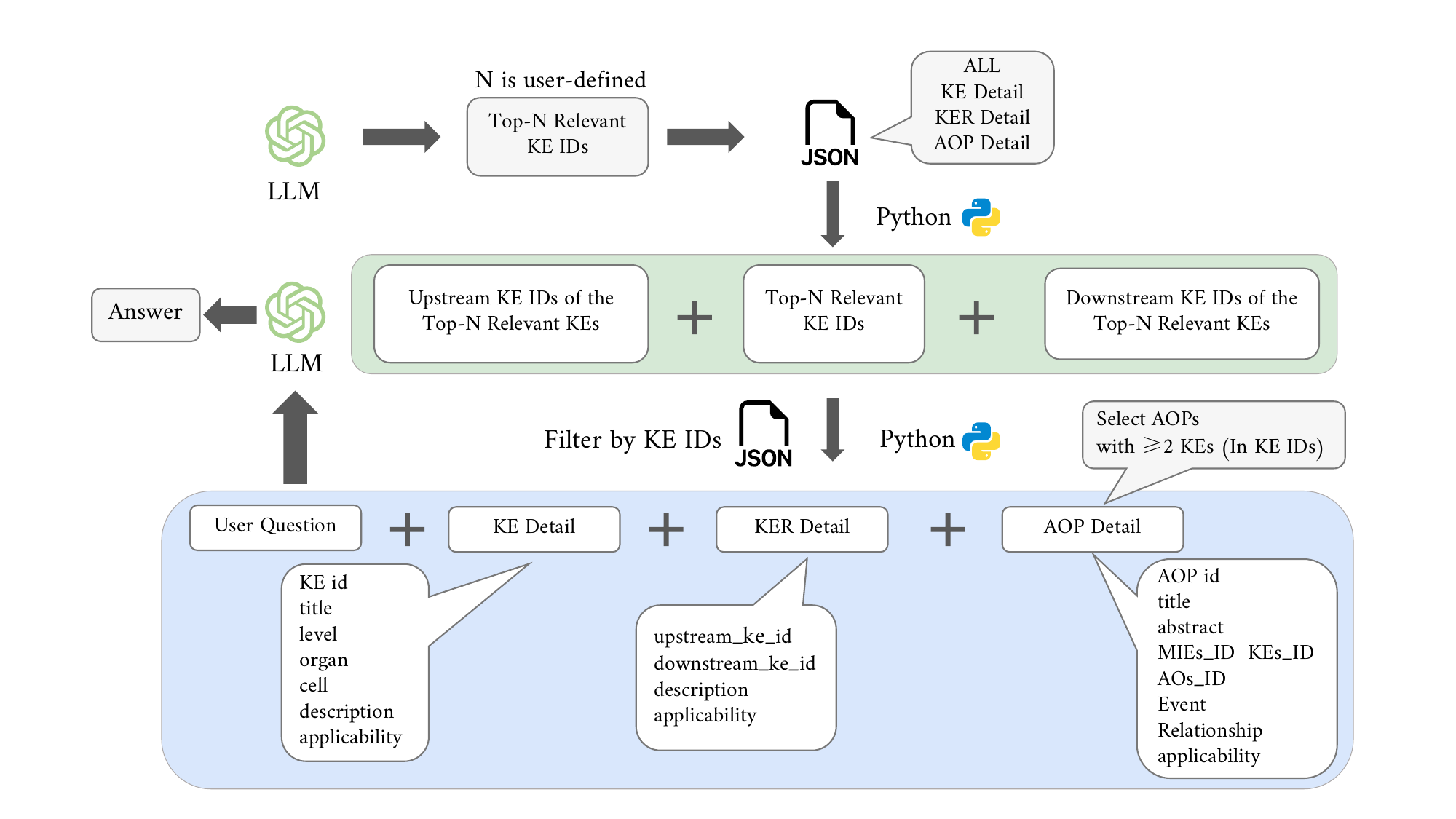}
    \caption{Overview of the KE-based expansion process, including KE augmentation, KER reconstruction, and AOP retrieval, followed by integration with LLM-based reasoning.}
    \label{fig:expansion}
\end{figure}

After the LLM completes the preliminary screening of KEs, the system performs expansion based on the selected KEs to enhance the coverage of question-related knowledge. The related process is shown in Figure~\ref{fig:expansion}. The specific process is as follows.

For each selected KE, the system first extracts its directly upstream and directly downstream KE IDs from \texttt{AOP-Smart.json}, and merges them with the originally selected KE to form an expanded KE set. The purpose of this step is to supplement the local mechanistic information causally adjacent to the events explicitly involved in the question.

After obtaining the expanded KE set, the system further traverses all Key Event Relationships (KERs) in the knowledge base, and selects those relationships that connect any two KEs within the expanded KE set, forming an expanded KER set. Through this joint expansion process, the system can not only identify the event nodes most relevant to the question, but also recover the causal relationships among these events, providing a basis for mechanism explanation and relationship judgment in subsequent responses.

Subsequently, the system scans all AOPs. When at least two matches exist between the MIE, KE, or AO in an AOP and the expanded KE set, the system includes that AOP into the expanded AOP set. This matching rule is intended to reduce the noise caused by accidental matching of a single event and improve the relevance between the retrieval results and the question. Through this mechanism, the system can further expand from local event relationships to complete AOP chains, thereby providing higher-level background knowledge support for subsequent question answering.

Finally, the system integrates the expanded KE, KER, and AOP information, and inputs them together with the user question into the LLM for subsequent reasoning and answer generation. In this way, the final output of the model is established on the basis of explicit external knowledge support, which helps reduce free generation detached from the facts in the knowledge base.

\subsection{Engineering Implementation}

\begin{figure}[t]
    \centering
    \includegraphics[width=0.9\columnwidth]{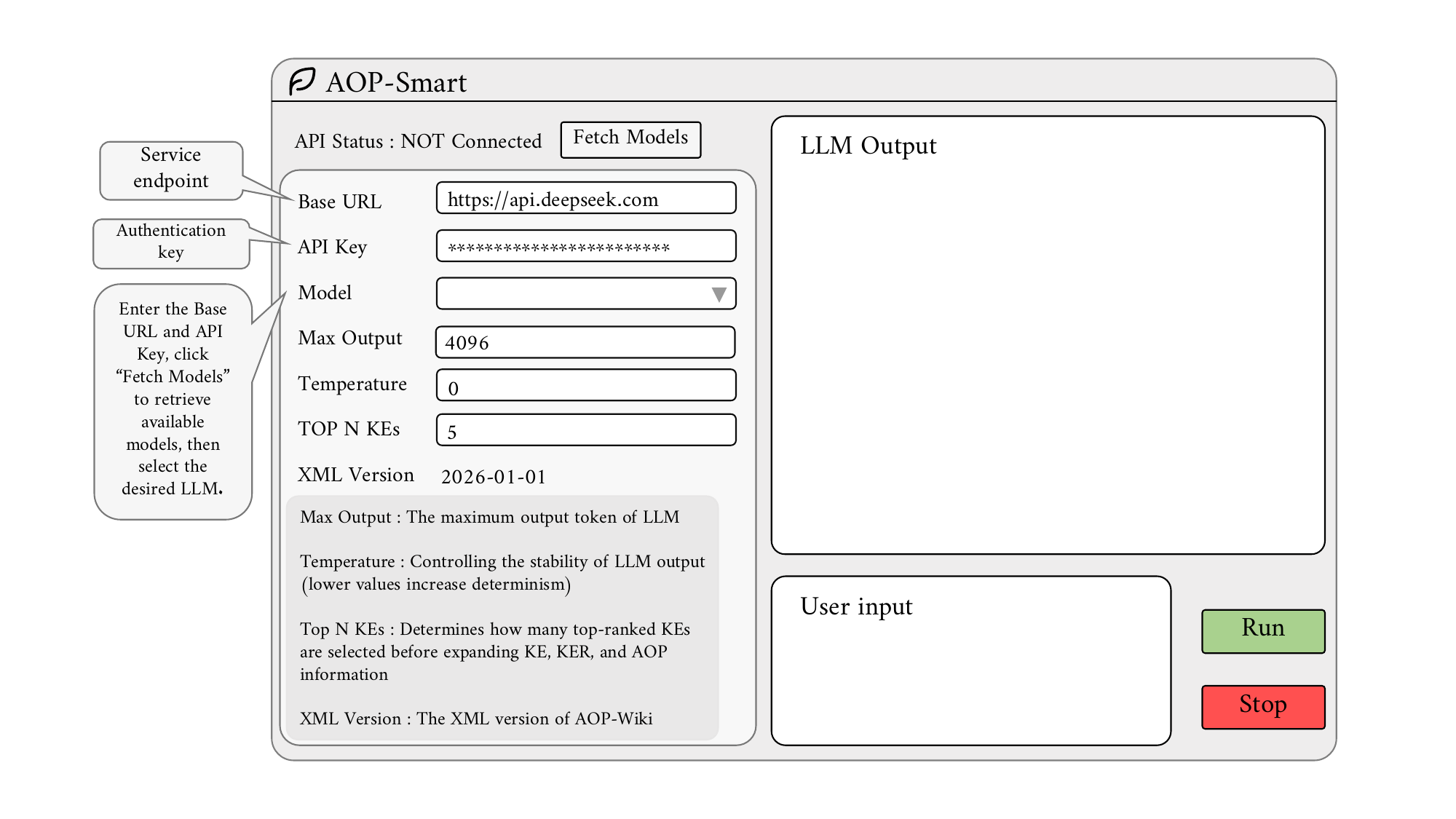}
    \caption{Overview of the Python-based system interface, including configuration, input, and output modules, with adjustable parameters for model selection and KE-based retrieval control.}
    \label{fig:system}
\end{figure}

This study implemented a complete software system based on Python, and its interface is shown in Figure~\ref{fig:system}.

The system mainly consists of three functional areas: the configuration area, the user input area, and the LLM output area.

In the configuration area, the user needs to input the Service Endpoint and API key of the large model being used. After completing the configuration, clicking the Fetch Models button will obtain the list of currently available large language models, and system initialization is completed by selecting a model.

The system also provides adjustable interfaces for multiple key parameters to support flexible control under different task requirements. These include: \texttt{Max Output}, which is used to control the maximum number of tokens generated by the model; \texttt{Temperature}, which is used to adjust the randomness and creativity of the model output, where a lower value leads to more stable output and a higher value leads to more diverse generated results; and the \texttt{Top-N KE} parameter, which is the core control variable in this study, used to specify the number of Key Events (KEs) screened by the LLM, thereby controlling the expansion scale and information coverage of RAG retrieval.

In the user input area, after the user enters the question to be processed and clicks the RUN button, the system will call the LLM based on the configuration parameters and the expanded knowledge retrieval mechanism, and output the final result to the output area.

The software system developed in this study and the related experimental code have been made public as supplementary materials to support method reproduction. The project code can be accessed via GitHub:
\url{https://github.com/qinjiang-lab/AOP-Smart}

\section{Experimental Design and Results}

\subsection{Experimental Design}

To evaluate the performance of large language models on AOP knowledge tasks and verify the effectiveness of the proposed method, this paper constructed an AOP question answering test set containing 20 questions, as shown in Figure~\ref{fig:benchmark}.The full set of questions is provided in the Supplementary Material.

\begin{figure}[htbp]
    \centering
    \includegraphics[width=0.9\columnwidth]{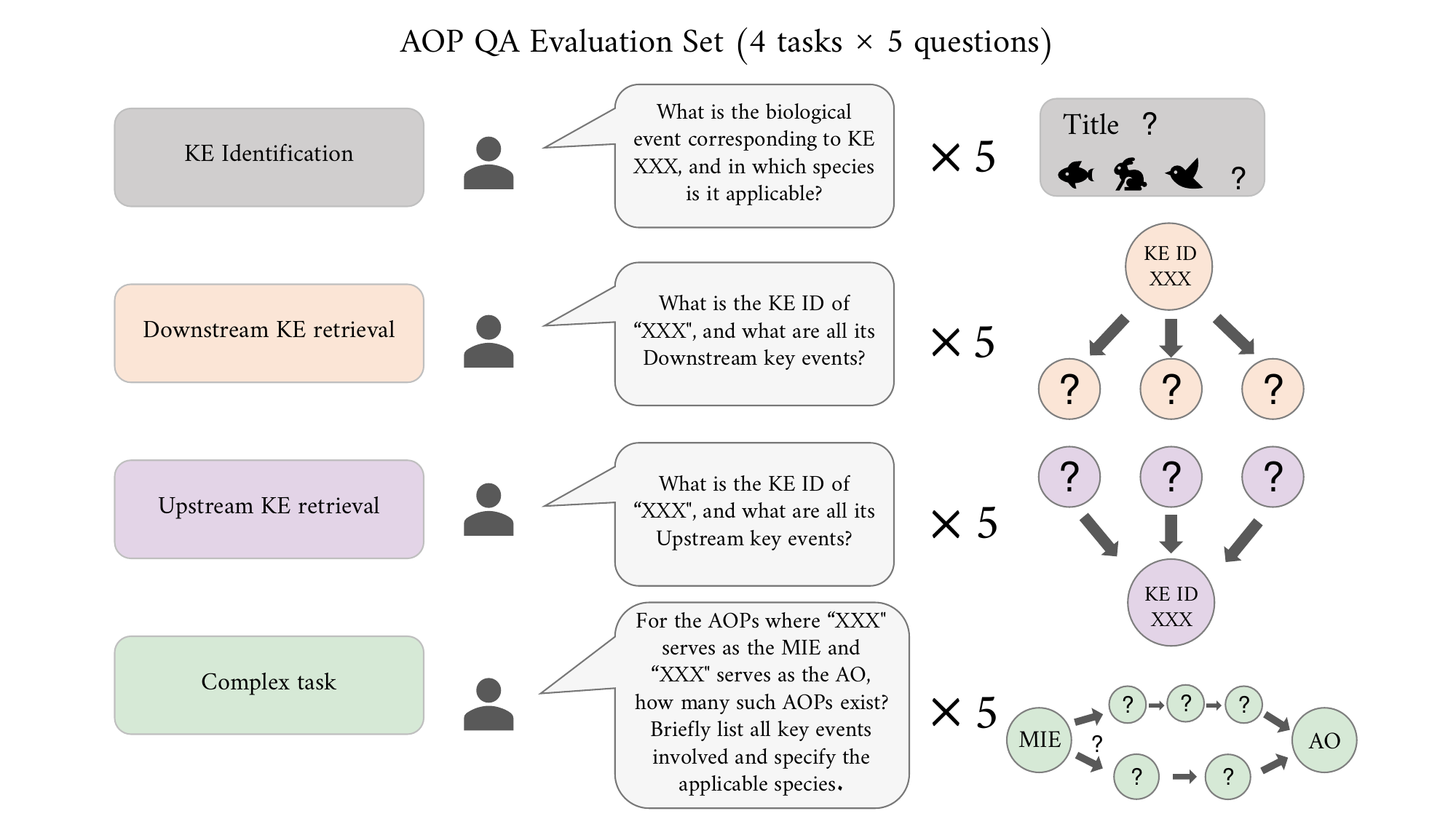}
    \caption{Overview of the 20-question AOP benchmark, grouped into four task categories: KE identification, downstream KE retrieval, upstream KE retrieval, and complex AOP query.}
    \label{fig:benchmark}
\end{figure}

According to different task types, the test set is divided into four categories:

\begin{enumerate}[label=(\arabic*)]
    \item \textbf{Key Event Identification (KE Identification):} Given a KE ID, the model is required to identify the corresponding biological event and its applicable species;
    \item \textbf{Downstream Key Event Retrieval (Downstream KE Retrieval):} Given an event name, the model is required to return the corresponding KE ID and list all directly downstream KEs;
    \item \textbf{Upstream Key Event Retrieval (Upstream KE Retrieval):} Given an event name, the model is required to return the corresponding KE ID and list all directly upstream Key Events;
    \item \textbf{Complex AOP Query Task (Complex AOP Query):} This involves retrieving the complete AOP chain based on a given MIE and AO, and requires listing the related Key Events and their applicable species information.
\end{enumerate}

The above tasks cover different types in AOP knowledge, such as KE identification, KER retrieval, and complex AOP query, and can evaluate the hallucination problem of LLMs from multiple perspectives.

The evaluation metric adopts Accuracy, defined as:
\[
\mathrm{Accuracy} = \frac{N_{\mathrm{correct}}}{N_{\mathrm{total}}}
\]
where $N_{\mathrm{correct}}$ represents the number of completely correct answered questions, and $N_{\mathrm{total}}$ represents the total number of questions. In this study, the factual information in the corresponding entries of AOP-Wiki is used as the basis for judgment, and when the model answer is completely consistent with the standard knowledge, it is judged as correct.

\subsection{Experimental Results}

In this experiment, the Top-N parameter was uniformly set to 5, and Temperature was set to 0 to ensure the retrieval scale and the determinism of the generated results. The specific models used were DeepSeek-V3.2, GPT-5.4, and Gemini-3.1 Pro, respectively. The experimental results are shown in Table~\ref{tab:results}.

\begin{table*}[t]
\centering
\small
\setlength{\tabcolsep}{9pt}

\caption{Experimental results of different large language models on AOP-related tasks with and without RAG enhancement.}
\label{tab:results}

\begin{tabular}{llcccccccc}
\toprule

Task & Description 
& \multicolumn{2}{c}{DeepSeek} 
& \multicolumn{2}{c}{Gemini} 
& \multicolumn{2}{c}{GPT} 
& \multicolumn{2}{c}{Overall} \\

\cmidrule(lr){3-4}
\cmidrule(lr){5-6}
\cmidrule(lr){7-8}
\cmidrule(lr){9-10}

& 
& \makecell{\footnotesize W/O\\RAG} & \makecell{\footnotesize W/\\RAG}
& \makecell{\footnotesize W/O\\RAG} & \makecell{\footnotesize W/\\RAG}
& \makecell{\footnotesize W/O\\RAG} & \makecell{\footnotesize W/\\RAG}
& \makecell{\footnotesize W/O\\RAG} & \makecell{\footnotesize W/\\RAG} \\

\midrule

Task 1 & KE Identification 
& 80\% & 100\% 
& 60\% & 100\% 
& 60\% & 100\% 
& 66.67\% & 100\% \\

Task 2 & Downstream KE retrieval 
& 20\% & 100\% 
& 20\% & 100\% 
& 0\% & 100\% 
& 13.33\% & 100\% \\

Task 3 & Upstream KE retrieval 
& 20\% & 100\% 
& 0\% & 80\% 
& 0\% & 100\% 
& 6.67\% & 93.33\% \\

Task 4 & Complex task 
& 20\% & 100\% 
& 0\% & 100\% 
& 0\% & 80\% 
& 6.67\% & 93.33\% \\

Overall & All tasks combined 
& 35\% & 100\% 
& 20\% & 95\% 
& 15\% & 95\% 
& 23.33\% & 96.67\% \\

\bottomrule
\end{tabular}
\end{table*}

Without introducing RAG, the performance of the three large language models on AOP-related question answering tasks showed obvious differences. Among them, the overall accuracies of DeepSeek, Gemini, and GPT were 35.0\%, 20.0\%, and 15.0\%, respectively, indicating that it is difficult to stably complete AOP tasks by relying only on the internal knowledge of the models.

After introducing the AOP-Smart retrieval-augmented framework proposed in this paper, the performance of each model on all tasks was significantly improved. The overall accuracy of DeepSeek increased to 100\% after enhancement, while Gemini and GPT increased to 95.0\% and 95.0\%, respectively, verifying the effectiveness of RAG in improving the ability of models to utilize external knowledge.

From the task-level analysis, RAG has a consistent improvement effect on different types of tasks. In the Key Event Identification (Task 1) and Downstream Key Event Retrieval (Task 2) tasks, all three models achieved accuracies close to or reaching 100\%; in the Upstream Key Event Retrieval (Task 3) and Complex AOP reasoning task (Task 4), RAG also significantly improved the performance of the models in multi-hop structural relationship queries, but compared with basic retrieval tasks, there were still certain differences in difficulty.

These results indicate that the RAG method can effectively improve the stability and reliability of large language models in professional toxicological knowledge question answering, and significantly alleviate the hallucination problem of LLMs.

\section{Limitations}

Although the AOP-Smart framework proposed in this study has achieved relatively significant experimental results in AOP-related tasks, there are still certain limitations.

First, the current method has not yet introduced information such as evidence level, evidence strength, and confidence in the AOP knowledge system. Therefore, in complex reasoning processes, it is not possible to perform weighted processing on knowledge from different sources, which may affect the reliability and interpretability of the final results.

Second, when constructing the retrieval knowledge base, this paper performed necessary text filtering and truncation processing on the KE, KER, and AOP descriptive information in AOP-Wiki to adapt to the input length limitation of large language models. Although this strategy improves computational efficiency, it may also lead to the loss of some contextual semantic information, thereby affecting the completeness of answers in complex query tasks.

Third, the evaluation set constructed in this study is relatively limited in scale (20 questions), and is mainly based on manually designed query tasks. Although it can cover typical KE identification and AOP relationship retrieval scenarios, it is still insufficient in covering larger-scale, multi-hop reasoning, and open-domain AOP question answering tasks.

Finally, different large language models are somewhat sensitive to prompt format and context organization methods. Therefore, the experimental results may be jointly affected by both model capability and prompt engineering design, and this factor has not yet been systematically analyzed through ablation in this study.

\section{Conclusion and Future Work}

This study proposes a retrieval-augmented generation framework for Adverse Outcome Pathway (AOP) analysis, AOP-Smart. This method constructs a structured knowledge base based on the official XML data of AOP-Wiki, and through an expansion mechanism based on Key Events (KEs) and Key Event Relationships (KERs), introduces AOP hierarchical structural information into the reasoning process of large language models, thereby enhancing their knowledge acquisition capability in professional toxicological question answering tasks.

The experimental results show that, on 20 structured AOP question answering tasks, AOP-Smart significantly improves the overall performance of DeepSeek, GPT, and Gemini under the no-RAG condition, and achieves near-optimal or optimal accuracy in most task types. Among them, in the complex AOP query task, the RAG mechanism has a clear enhancement effect on multi-hop structural relationship reasoning.

The above results indicate that the retrieval-augmented method based on structured AOP knowledge can effectively improve the stability and accuracy of large language models in professional-domain question answering tasks, and alleviate the model hallucination problem to a certain extent.

Future work will be carried out from the following three aspects. First, expand the scale of evaluation data and construct a larger-scale and multi-type AOP question answering benchmark to improve the comprehensiveness and robustness of evaluation. Second, introduce structured information such as evidence level and confidence, and perform weighted modeling on retrieval results to enhance the interpretability of the reasoning process. Finally, explore the application potential of AOP-Smart in automatic information extraction from scientific literature and AOP knowledge graph completion, so as to provide support for the automated construction and expansion of the AOP system.

\bibliographystyle{unsrt}
\bibliography{references}

\end{document}